\documentclass[conference]{IEEEtran}
\IEEEoverridecommandlockouts

\usepackage{cite}
\usepackage{amsmath,amssymb,amsfonts}
\usepackage{algorithmic}
\usepackage{graphicx}
\usepackage{textcomp}
\usepackage{xcolor}
\def\BibTeX{{\rm B\kern-.05em{\sc i\kern-.025em b}\kern-.08em
    T\kern-.1667em\lower.7ex\hbox{E}\kern-.125emX}}
\begin{document}

\title{Automated Road Extraction from Satellite Imagery Integrating Dense Depthwise Dilated Separable Spatial Pyramid Pooling with DeepLabV3+}

\author{
\IEEEauthorblockN{Arpan Mahara\IEEEauthorrefmark{1},
Md Rezaul Karim Khan \IEEEauthorrefmark{2}, Liangdong Deng\IEEEauthorrefmark{3},
Naphtali D. Rishe\IEEEauthorrefmark{4},
Wenjia Wang\IEEEauthorrefmark{5},
Seyed Masoud Sadjadi\IEEEauthorrefmark{6}}
\IEEEauthorblockA{Knight Foundation School of Computing and Information Sciences\\
Florida International University\\
Miami, FL 33199\\
\IEEEauthorrefmark{1}amaha038@fiu.edu,
\IEEEauthorrefmark{2}mkhan157@fiu.edu,
\IEEEauthorrefmark{3}liadeng@cs.fiu.edu,
\IEEEauthorrefmark{4}rishen@cs.fiu.edu,
\IEEEauthorrefmark{5}wwang048@fiu.edu,
\IEEEauthorrefmark{6}sadjadi@cs.fiu.edu
}
}

\maketitle

\begin{abstract}
Road Extraction is a sub-domain of Remote Sensing applications; it is a subject of extensive and ongoing research. The procedure of automatically extracting roads from satellite imagery encounters significant challenges due to the multi-scale and diverse structures of roads; improvement in this field is needed. The DeepLab series, known for its proficiency in semantic segmentation due to its efficiency in interpreting multi-scale objects’ features, addresses some of these challenges caused by the varying nature of roads. The present work proposes the utilization of DeepLabV3+, the latest version of the DeepLab series, by introducing an innovative Dense Depthwise Dilated Separable Spatial Pyramid Pooling (DenseDDSSPP) module and integrating it in place of the conventional Atrous Spatial Pyramid Pooling (ASPP) module. This modification enhances the extraction of complex road structures from satellite images. This study hypothesizes that the integration of DenseDDSSPP, combined with an appropriately selected backbone network and a Squeeze-and-Excitation block, will generate an efficient dense feature map by focusing on relevant features, leading to more precise and accurate road extraction from Remote Sensing images. The results section presents a comparison of our model’s performance against state-of-the-art models, demonstrating better results that highlight the effectiveness and success of the proposed approach.
\end{abstract}

\begin{IEEEkeywords}
DeepLabV3+, ASPP, deep learning, semantic segmentation, satellite imagery, Xception, remote sensing, Squeeze-and-Excitation, road extraction.
\end{IEEEkeywords}

\section{Introduction}
The availability of high-resolution satellite images and development of methodologies to extract roads from satellite images have revolutionized the Remote Sensing domain. Some sub-domains predominately impacted by this revolution include autonomous navigation, transportation management, urban development planning, and so on. However, the diverse variations in road structures introduce multi-scale characteristics, which in turn lead to limitations in accurate road extraction. Additionally, the sparsity of roads and the presence of shadows in Remote Sensing images present significant challenges in the automatic and systematic extraction of roads from satellite imagery. 

Mnih and Hinton's \cite{mnih2010learning} study presents the pioneering work in automatic road detection from high-resolution satellite imagery on a large scale. They effectively incorporated a neural network with millions of trainable weights and successfully equipped the network to detect road objects in large datasets of urban imagery automatically. Following the work of Mnih, several deep learning methods have been utilized to automate road detection and extraction from Remote Sensing images. The methods range from general deep learning algorithms \cite{langkvist2016classification}, \cite{sherrah2016fully}, \cite{mahara2024dawn} such as fully-convolution networks to task-specific models \cite{8309343}, \cite{mahara2024multispectral}, \cite{gao2018end}. 

While these advanced deep learning methods were successful in the semantic segmentation task, such as extracting roads from satellite imagery, the downsampling operation in the convolution layers may lead to the loss of spatial information \cite{xu2023haar}, a challenge also noted in feature selection for DDoS detection models \cite{wang2022curse} and in privacy-preserving federated learning approaches for medical image analysis \cite{das2023privacy}. To address this problem in semantic segmentation, atrous convolution was utilized in the study by Chen et al. \cite{chen2017deeplab}, which allows for learning of semantic information without spatial loss. This work also proposed an atrous spatial pyramid pooling (ASPP) module that performed better in capturing multi-scale objects. This approach was expanded in DeepLabV3 \cite{chen2017rethinking}, which can capture global context, followed by DeepLabV3+ \cite{chen2018encoder}, which incorporates a decoder module, with each progression showing improved performance in the semantic segmentation task. Another type of convolution called Depthwise separable convolution if used efficiently can lead to better performance as showcased by Chollet et al. \cite{chollet2017xception} by using them in their proposed model called Xception achieving state-of-the-art performance in classification task.

Due to their relevance and effectiveness in multi-scale feature capture, the DeepLab series has been explored for road extraction in recent literature \cite{quan2021improved}, \cite{wang2022road}, \cite{linghu2023information}. These studies mainly focus on altering the backbone feature extractor in the decoder, altering the loss functions, or incorporating new backbone extractors, such as VGG19 \cite{simonyan2014very}, ResNet50 \cite{he2016deep}, and Xception \cite{chollet2017xception}, alongside the same ASPP module. Even though the ASPP module has successfully captured multi-scale patterns, it may not adequately capture complex multi-scale features in objects such as roads. The DenseASPP \cite{yang2018denseaspp}, proposed by Yang et al., has effectively captured complex patterns in street scenes due to its capability to interpret dense features. However, despite its effectiveness, DenseASPP tends to be computationally intensive. Considering the contributions and limitations of these studies, our contributions to road extraction are as follows:

\begin{itemize}
\item We propose an innovative module called Dense Depthwise Dilated Separable Spatial Pyramid Pooling (DenseDDSSPP) and replace the ASPP module with it in DeepLabV3+.
\item We conducted an experimental evaluation of various deep learning models to identify an optimal backbone network, which is Xception.
\item The present work integrates the Squeeze-and-Excitation block in the decoder to enable our road extraction process to focus on relevant feature channels from the dense feature map obtained.
\item Our study demonstrates better performance of our proposed model in road extraction compared to state-of-the-art methods across different comparison metrics in a supervised setup.
\end{itemize}

\section{Related Work}
Applications of deep learning models have favored the advancement in road extraction from satellite imagery. Mnih and Hinton's \cite{mnih2010learning} pioneering work in neural network-based road extraction laid the foundation for subsequent advancements. Their model demonstrated the potential of large neural networks in handling extensive datasets for road detection. 

U-Net \cite{ronneberger2015u}, an advanced model well-known for its efficacy in semantic segmentation, has been well-adapted with extensive application in road extraction in recent literature \cite{zhang2018road} \cite{hou2021c}, \cite{yang2022sdunet}, \cite{akhtarmanesh2023road}. Zhang et al. \cite{zhang2018road} introduced Deep Residual U-Net by incorporating residual units into the U-Net architecture, achieving better performance in road extraction over its predecessors, including the original U-Net \cite{ronneberger2015u} and Mnih-CNN \cite{mnih2010learning}. Similarly, SegNet \cite{badrinarayanan2017segnet}, which utilizes an encoder-decoder framework with an encoder architecture topologically similar to VGG16 \cite{simonyan2014very}, was developed to create a computationally efficient segmentation model. SegNet model outperformed previous models in understanding and interpreting road scenes. 

Continuing with the chronological advancements of U-Net, Akhtarmanesh et al. \cite{akhtarmanesh2023road} integrated techniques such as a patch-based attention mechanism and rotation-based augmentation into the original U-Net. This advanced attention U-Net outperformed prior studies, including Residual U-Net \cite{zhang2018road}, SDUNet \cite{yang2022sdunet}, and SegNet \cite{badrinarayanan2017segnet}, based on Precision and IOU scores.

Deep Convolutional Neural Networks (DCNNs) often face reduced spatial resolution and challenges in interpreting objects at multiple scales, leading to degradation in semantic segmentation performance. To address these limitations, DeeplabV3 \cite{chen2017rethinking} employs atrous convolution within the ASPP module, achieving better results through a set of parallel atrous convolution determined by the output stride. DeepLabV3+ \cite{chen2018encoder}, an extension to DeepLabV3, incorporates a decoder module, further enhancing segmentation accuracy at object boundaries. Quan et al. \cite{quan2021improved} proposed an improvement to the DeepLabV3 model by fusing U-Net and utilizing both DICE and BCE losses during model training. This study addressed class imbalance in two-class samples, achieving effective road extraction from Remote Sensing images. 
Similarly, Linghu et al. \cite{linghu2023information} introduced improvements to DeepLabV3+ by incorporating MobileNetV2 as a backbone feature extractor and employing the Dice Loss function, achieving higher overall accuracy while reducing the model’s parameters for decreased computational complexity. Despite these advancements, modifications to the core architecture, specifically the ASPP module, remained unexplored in the road extraction domain. Wu et al. \cite{wu2020automatic} proposed a dense and global spatial pyramid pooling (DGSPP) module, inspired by the ASPP module, but did not take advantage of the encoder-decoder architecture of DeepLabV3+.
Building on the contributions and addressing the limitations of previous studies, this study proposes advancements in road extraction by replacing the ASPP module with DenseDDSSPP in the DeepLabV3+ model \cite{chen2018encoder}. In this approach, the output from a preceding depthwise separable convolution layer, after applying a dilation rate, is merged with the input of the next layer in an iterative procedure. Additionally, our study incorporates Squeeze-and-Excitation \cite{hu2018squeeze} in the decoder module to support the selection of relevant features for road extraction decisions. We hypothesize that this approach will efficiently generate denser features that capture intricate and useful road patterns, potentially leading to improved road extraction from high-resolution satellite imagery.

\section{Proposed Approach}
In this section, we revisit the foundational concepts of atrous convolution, depthwise dilated separable convolution, and ASPP before detailing the integration of DenseDDSSPP into the DeepLabV3+ architecture. This integration intends to enhance road extraction performance beyond current state-of-the-art models by leveraging the proficiency of dense feature extraction at multiple scales.

\subsection{Dilated Convolution in Spatial Pyramid Pooling}
Dilated convolution, also called atrous convolution, introduces a dilation rate as a parameter to conventional convolution operations \cite{chen2014semantic}, enabling accurate segmentation. This technique allows convolutional filters to expand their receptive field without altering the feature map resolution or increasing the computational cost. Mathematically, the dilated convolution equation is illustrated as:

\begin{equation}
Y(f) = \sum_{s=1}^{S} X(f + d \cdot s)\cdot W(s)
\end{equation}

where \(X[f]\) is the input feature map, \(W(s)\) is the s-th weight in the filter, \(S\) denotes the filter's size, and \(d\) denotes the dilation rate. \(Y[f]\) is the resultant output feature map. Increasing \(d\) enlarges the receptive field, improving the model's pixel-level classification capabilities by capturing broader contextual information. Dilated convolution achieves this improvement by convolving the input \(X\) with a filter modified to insert \(d-1\) zeros between consecutive filter values, expanding the filter's coverage area without increasing the number of parameters. This approach prevents the spatial resolution loss commonly associated with the downsampling operations in conventional convolution and helps infer a larger field of view for making road extraction decisions without increasing computation.

\begin{figure}[tb]
  \centering
  \includegraphics[height=6cm]{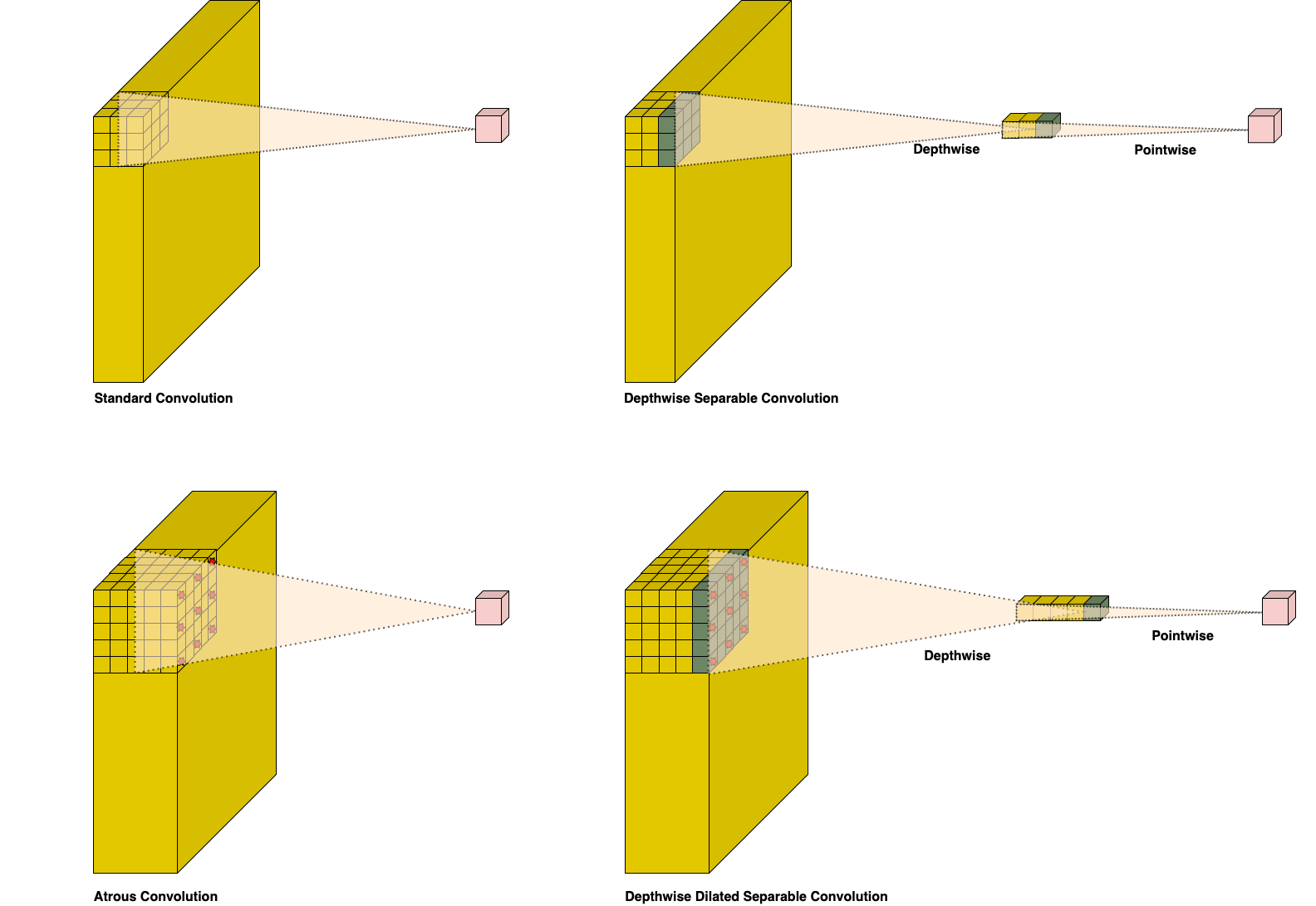}
  \caption{Depiction of Different Convolutions}
  \label{fig:DifferentConvolutions}
\end{figure}

\subsection{Depth-wise Dilated Separable Convolution}
Depth-wise dilated separable convolution expands the effectiveness of dilated convolution by combining depthwise separability with a dilation rate, enhancing computational efficiency and receptive field coverage (as depicted in Fig. \ref{fig:DifferentConvolutions}). This method involves two steps: a) performing depth-wise dilated convolution on each input channel independently with a dilation rate of \(d\), increasing the receptive field to \(r_d \times r_d\), and b) applying a point-wise convolution to learn linear combinations of the depth-wise convolution outputs \cite{mehta2019espnetv2}. Here, \(r \times r\) is the receptive field of a regular convolution for learning representation.

Mathematically, we represent depth-wise dilated separable convolution as:

\begin{equation}
Y(f, k) = \sum_{c=1}^{C} \left( \sum_{s=1}^{S} X_c(f + d \cdot s) \cdot W_{c,k}(s) \right)
\end{equation}

where \(X_c[f]\) is the input feature map at channel \(c\), \(W_{c,k}(s)\) is the s-th weight in the depth-wise filter for channel \(c\) and output channel \(k\), \(S\) denotes the filter's size, and \(d\) denotes the dilation rate. The inner summation represents the depth-wise dilated convolution, and the outer summation represents the point-wise convolution across all input channels \(C\). \(Y(f, k)\) is the resultant output feature map for the k-th output channel.

This convolution reduces computational complexity while maintaining an increased field-of-view for feature representation learning, making it a potentially efficient and powerful tool for road extraction from satellite imagery.

\subsection{ASPP Module}
The advent of atrous convolution mitigated the issue of spatial resolution loss, but the appearance of variable-sized objects in images introduced new challenges for accurate semantic segmentation. To overcome these challenges, Chen et al. \cite{chen2017rethinking} proposed using atrous convolution in a cascading or parallel fashion within the ASPP module. In the cascading mechanism of the ASPP module, the output from a lower atrous layer is passed to a higher layer, producing larger receptive fields. Meanwhile, parallel processing in the ASPP module involves feeding the same input to multiple atrous layers with varying dilation rates. The output obtained from each layer are concatenated to form a comprehensive feature map. This feature map now contains the information of the input across different scales. Mathematically, atrous convolution with $H_{K,d}(x)$, and ASPP can be illustrated as:
\begin{equation}
y = H_{3,6}(x) + H_{3,12}(x) + H_{3,18}(x) + H_{3,24}(x)
\end{equation}

The value of dilation rates {6, 12, 18, 24} was based on the output stride \cite{chen2017rethinking}. The multi-scale feature aggregation utilizing atrous convolution at various dilation rates in ASPP is a key factor in improving road extraction, particularly at object boundaries.

\subsection{DeepLabV3+}

DeepLabV3+ \cite{chen2018encoder}, an advancement in the DeepLab series, proposes an encoder-decoder architecture. The encoder employs an ASPP module after processing the input via backbone networks such as VGG19 \cite{simonyan2014very}, ResNet50 \cite{he2016deep}, Xception \cite{chollet2017xception}, and so on for a refined feature understanding. Post-ASPP processing, the decoder concatenates the resultant feature with features sourced from the initial stages of the same backbone network. This concatenation can recover object boundaries, leading to better outcomes in semantic segmentation.

\begin{figure*}[t]
  \centering
  \includegraphics[width=1\linewidth]{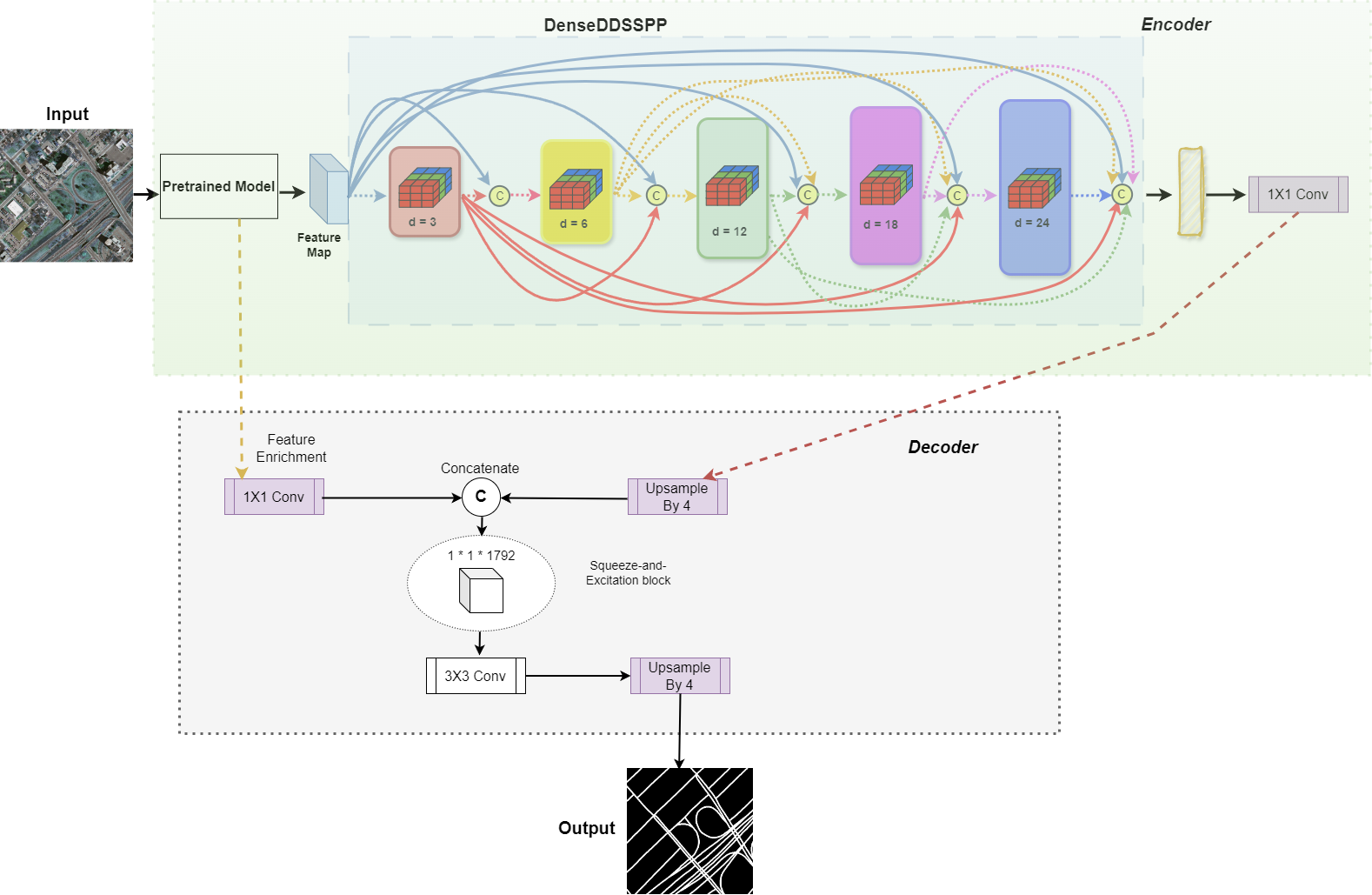}
  \caption{Architecture of DeepLabV3+ with DenseDDSSPP module}
  \label{fig:DeepLabV3pluswithDenseDDSSPP}
\end{figure*}

\subsection{Integration of DenseDDSSPP into the Network}
We propose the integration of DenseDDSSPP into the DeepLabV3+ architecture, replacing the standard ASPP module to achieve more accurate road extraction from satellite imagery (as depicted in Fig. \ref{fig:DeepLabV3pluswithDenseDDSSPP}). Motivated by the work of Yang et al. \cite{yang2018denseaspp}, we arrange depth-wise dilated separable convolution layers in a cascade, increasing the dilation rate of layers in ascending order. Unlike ASPP, the output of any momentary layer obtained by applying depth-wise dilated separable convolution with a selected dilation rate is concatenated with both the input feature map and the output feature maps from all previous layers, generating an efficient dense feature map. The resultant concatenated feature map is then fed to the following layer. The present study performs these operations for all layers in the DenseDDSSPP module iteratively to obtain the final dense feature map (as shown in Fig. \ref{fig:DeepLabV3pluswithDenseDDSSPP}, in the encoder section with different colored arrows). In the present study, the computational efficiency in DenseDDSSPP is enhanced by using depthwise separable convolutions with dilation rates instead of traditional convolutions. This approach involves a depthwise convolution with a dilation rate followed by a pointwise convolution, reducing computational complexity while maintaining the ability to capture multi-scale features. Tracing back to equation (2) and (3), this layered approach in DenseDDSSPP is mathematically illustrated as follows:

\begin{equation}
Y_l = D_{S,d_l}([Y_{l-1}, Y_{l-2}, \ldots, Y_0])
\end{equation}
where \( l \) denotes the layer index, \( d_l \) is the dilation rate for layer \( l \), and concatenation is denoted by \([ \ldots ]\). The expression \([Y_{l-1}, \ldots, Y_0]\) denotes the feature map resulted by merging the outputs from all preceding layers. Here, \( D_{S,d_l} \) represents the depth-wise dilated separable convolution operation with kernel size \( S \) and dilation rate \( d_l \). These above modifications and integration provides two explicit advantages i.e., efficient denser feature map and larger receptive field, which are crucial for accurately interpreting complex road structures.

\begin{figure}[tb]
  \centering
  \includegraphics[width=1\linewidth]{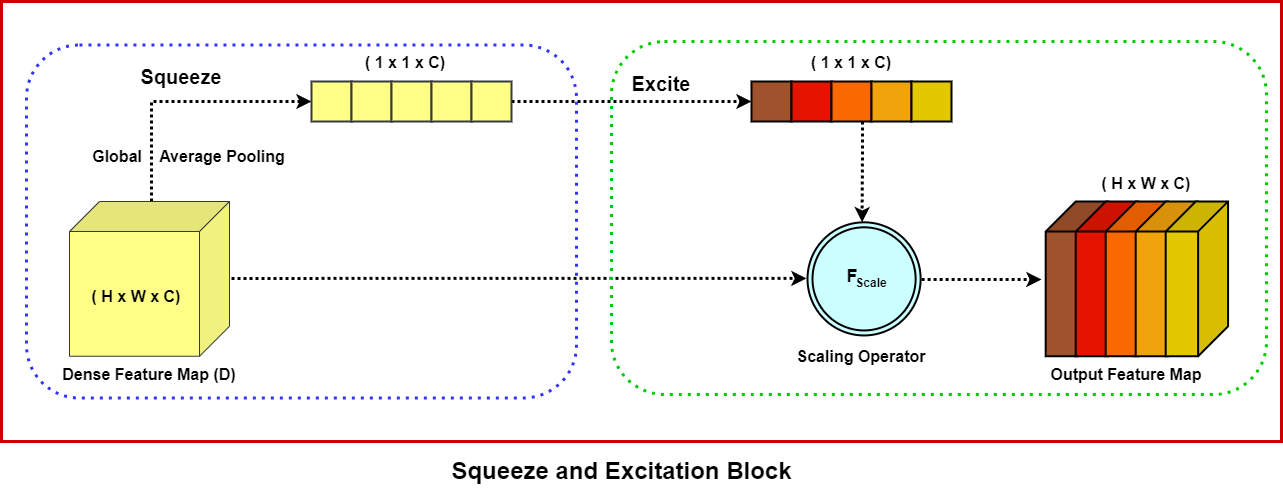}
  \caption{Squeeze-and-Excitation Block}
  \label{fig:SqueezeandExcitationBlock}
\end{figure}

\subsubsection{Selection of suitable backbone network and Incorporation of Squeeze-and-Excitation (SE) block:}

Since selection of relevant backbone network is crucial in the DeepLabV3+ architecture, the present study made an experimental evaluation on different networks to determine the suitable one. After careful evaluation, Xception was selected as the ultimate backbone network in the present work. We utilize the \(block3\_sepconv2\) layer of Xception for low-level feature extraction. Similarly, we selected the \(block13\_sepconv2\_bn\) layer for high-level features extraction, which is then fed into the DenseDDSSPP module. Consequently, the process following this aligns with the latest DeepLab series, i.e., DeepLabV3+ \cite{chen2018encoder} by concatenating dense feature maps with low-level features at the decoder.

In our architecture, after concatenation at the decoder, the resulting features are further enhanced using the Squeeze-and-Excitation (SE) block \cite{hu2018squeeze}, a technique capable of effectively configuring channel-wise feature responses (as depicted in Fig. \ref{fig:SqueezeandExcitationBlock}).

The output dense feature map obtained from concatenation, denoted as \(D\) with \(C\) channels, where each channel \(c\) has height \(H\) and width \(W\), undergoes a \(squeeze\) operation. The \(squeeze\) operation can be mathematically expressed as:
\begin{equation}
z_c = F_{\text{sq}}(d_c) = \frac{1}{H \times W} \sum_{i=1}^{H} \sum_{j=1}^{W} d_c(i, j)
\end{equation}
This is the global average pooling operation.

To distinguish between the important channels that lead to better road extraction, we perform the \(excitation\) operation on the information obtained from the squeeze operation. Similar to the original work of Hu et al. \cite{hu2018squeeze}, which acts as an attention mechanism in a channel-wise manner, the \(excitation\) operation can be written as:
\begin{equation}
e = F_{\text{ex}}(z, W) = \sigma(W_2 \delta(W_1 z))
\end{equation}
where \(W_1 \in \mathbb{R}^{C/r \times C}\) is the reduction to dimension \(C/r\), \(W_2 \in \mathbb{R}^{C \times C/r}\) scales back to dimension \(C\), \(\sigma\) is the sigmoid activation, and \(\delta\) is the ReLU activation.

Finally, we obtain the final feature volume that depicts the importance of one channel over another by a scaling operation. Given the feature map \(D\) and the scalar \(e\), the scaling operation can be represented as:
\begin{equation}
x_c = F_{\text{scale}}(d_c, s_c) = s_c \cdot d_c
\end{equation}
where \(X = [x_1, x_2, \ldots, x_C]\) and \(F_{\text{scale}}\) refers to channel-wise multiplication.

This completes the enhancement of features through the Squeeze-and-Excitation block in our architecture.

The integration of this SE block post-decoder ensures that our model captures comprehensive spatial and contextual information, additionally supporting attention mechanism to focus more on relevant patterns for improved road extraction accuracy.
Finally, aligning to the original DeepLabV3+ \cite{chen2018encoder}, the above operation is followed by convolution operations and bilinear upsampling, concluding with a sigmoid activation function ensuring accurate road structures extraction from the satellite imagery.

\section{Experiments and Results}

\subsection{Datasets}

\subsubsection{Massachusetts Road Dataset:}

For the experiments and comparison, we used the Massachusetts Roads dataset, as collected and presented by Mnih and Hinton \cite{mnih2010learning}. This dataset covers roughly 2600 square kilometers of Massachusetts state, comprising 1171 satellite images along with their corresponding road-extracted mask images. Each image is \(1500 \times 1500\) pixels in dimensions, with a resolution of 1 meter per pixel. Given the high dimensions of these images, we employed a patchify procedure, cropping each image to obtain \(512 \times 512\) tiles. This step was crucial to reduce computational demands and enable the training of complex deep learning models. From the entire dataset, 817 images were selected for our experiments, resulting in 3268 images and corresponding masks of \(512 \times 512\) dimensions. For training and testing, the dataset was divided using an 80:20 split.

\subsubsection{DeepGlobe Road Dataset:}
Similarly, we used the DeepGlobe Dataset \cite{demir2018deepglobe} as second dataset to test the performance of our proposed method. It consists of 6,226 images with their corresponding road extracted masks, with a resolution of \(1024 \times 1024\). To maintain consistency while considering computational demands, we cropped the images and masks into \(512 \times 512\) tiles, resulting in 24,904 images and 24,904 masks. Due to memory constraints, we selected 9000 images and their corresponding masks and divided the dataset using an 80:20 split into training and testing sets.

\subsection{Evaluation Metrics}

In this study, we selected commonly used metrics such as Precision, \(F_1\) Score, and Intersection over Union (IOU) to assess the model's performance. These metrics are defined as follows:

Precision is the ratio of the number of correctly predicted positive observations to the total predicted positive observations:
\begin{equation}
\text{Precision} = \frac{TP}{TP + FP}
\end{equation}
where \(TP\) represents the true positives, and \(FP\) represents the false positives.

The \(F_1\) Score, which measures the accuracy of positive predictions, is the harmonic mean of Precision and Recall. It can be simplified and expressed mathematically as: 
\begin{equation}
F_1 = \frac{2TP}{2TP + FP + FN}
\end{equation}
where \(FN\) represents the false negatives.

Intersection Over Union (IOU), which evaluates the pixel-wise accuracy of the segmentation, can be mathematically illustrated as:
\begin{equation}
\text{IOU} = \frac{TP}{TP + FP + FN}
\end{equation}

Alternatively, IOU can be understood geometrically as the overlap between the predicted segmentation and the ground truth segmentation.

\subsection{Experimental Environment and Baselines}
We conducted our experiments using the Python 3.6.8 environment with TensorFlow framework for all aspects of training and testing. The computational tasks were carried out on a system comprising eight NVIDIA A100-PCI GPUs, each featuring 80 GB of HBM2 memory. The computation tasks were facilitated using CUDA version 11.8 and NVIDIA driver version 520.61.05. We utilized a mirrored-strategy and distributed the computation across 4 GPU machines to manage the intensive deep learning tasks. All the models were trained for 300 epochs using Adam as the optimizer, with an exponential decay procedure. The initial learning rate was set to 0.001, with the decay maintained at steps of 10,000 and a decay rate of 0.96 to ensure efficient training and better convergence of the models. 

\subsubsection{Baselines}
To validate the efficacy of our proposal, we selected supervised state-of-the-art models in the road-extraction domain. The selected models are U-Net \cite{ronneberger2015u}, Attention U-Net \cite{akhtarmanesh2023road}, DeepLabV3+ with ASPP \cite{chen2018encoder}, SegNet \cite{badrinarayanan2017segnet}, RFE-Link Net \cite{zhao2023rfe}, \cite{wu2020automatic}, and D-LinkNet \cite{zhou2018d}.  

\subsection{Comparison and Results}
The quantitative results obtained from different state-of-the-art models, including our proposed model, on the Massachusetts road dataset are presented in Table \ref{tab:model_comparison_massachusetts} and the DeepGlobe road dataset are presented in Table \ref{tab:model_comparison_deepglobe}. Models such as U-Net \cite{ronneberger2015u}, Attention U-Net \cite{akhtarmanesh2023road}, DeepLabV3+ with ASPP \cite{chen2018encoder}, and SegNet \cite{badrinarayanan2017segnet} were experimented within our setup, and the metrics' results were reported following training and testing. The results for RFE-Link Net \cite{zhao2023rfe} and D-LinkNet \cite{zhou2018d} and \cite{wu2020automatic} are reported from the original papers due to the relevancy in setup and usage of the same dataset. 
As seen in Table \ref{tab:model_comparison_massachusetts}, our proposed model achieved better performance in Massachusetts dataset based on IOU and Precision metrics, with scores of 67.21 and 81.38, respectively, compared to all other models. Additionally, our model showcased better performance compared to all models in \(F_1\) Score, with 79.29, while the advanced model RFE-Link Net slightly outperformed our model with a score of 80.07.

Similarly, as seen in Table \ref{tab:model_comparison_deepglobe}, our proposed model was also successful to achieve better performance in DeepGlobe dataset in IOU and Precision metrics, with scores of 71.61 and 83.19, respectively, compared to all other models. Along this, the experimental results follow the same trend as obtained in above, in which our model again depicted better performance compared to all models in \(F_1\) Score, with 81.75, while the advanced model RFE-Link Net slightly outperformed our model with a score of 82.85.

\begin{table}[h]
\centering
\caption{Quantitative Observation of Results Obtained from All Models in Massachusetts Road Dataset}
\label{tab:model_comparison_massachusetts}
\begin{tabular}{|l|c|c|c|c|}
\hline
Model & IOU (\%) & Precision (\%) & \(F_1\) Score (\%) \\ \hline
U-Net \cite{ronneberger2015u}              & 64.19 & 80.23 & 74.78 \\ \hline
Original DeepLabV3+ \cite{chen2018encoder} & 65.92 & 80.04 & 75.60 \\ \hline
SegNet \cite{badrinarayanan2017segnet}     & 58.67 & 78.56 & 73.73\\ \hline
\cite{wu2020automatic}                     & 62.48   & -     & 76.59 \\ \hline
D-LinkNet\cite{zhou2018d}                  & 63.74 & 75.89 & 77.86 \\ \hline
RFE-LinkNet \cite{zhao2023rfe}             & 66.77 & 80.88 & \textbf{80.07} \\ \hline
Proposed Model                             & \textbf{67.21} & \textbf{81.38} & 79.29\\ \hline
\end{tabular}
\end{table}

\begin{table}[h]
\centering
\caption{Quantitative Observation of Results Obtained from All Models in DeepGlobe Road Dataset}
\label{tab:model_comparison_deepglobe}
\begin{tabular}{|l|c|c|c|c|c|}
\hline
Model & IOU (\%)  & Precision (\%) & \(F_1\) Score (\%) \\ \hline
U-Net \cite{ronneberger2015u}            &62.82   & 80.36  &  71.83\\ \hline
DeepLabV3+ \cite{chen2018encoder}        &69.05   & 83.16  & 77.96 \\ \hline
SegNet \cite{badrinarayanan2017segnet}   &57.66   & 80.67  & 72.09 \\ \hline
Attention-UNet\cite{akhtarmanesh2023road}&67.42   & 79.95  & 80.54\\ \hline
D-LinkNet\cite{zhou2018d}                &63.94   & 78.54  & 0.7659 \\ \hline
RFE-Link Net \cite{zhao2023rfe}          &70.72   & 83.09  & \textbf{82.85}\\ \hline
Proposed Model                           & \textbf{71.61} & \textbf{83.19} & 81.75 \\ \hline
\end{tabular}
\end{table}

\begin{figure*}[!t]
    \centering
    \setlength{\tabcolsep}{3pt}
    \begin{tabular}{cccccc}
        Input & GroundTruth & Proposed & DeepLabV3+ & U-Net & SegNet\\
        
        \includegraphics[width=0.15\textwidth]{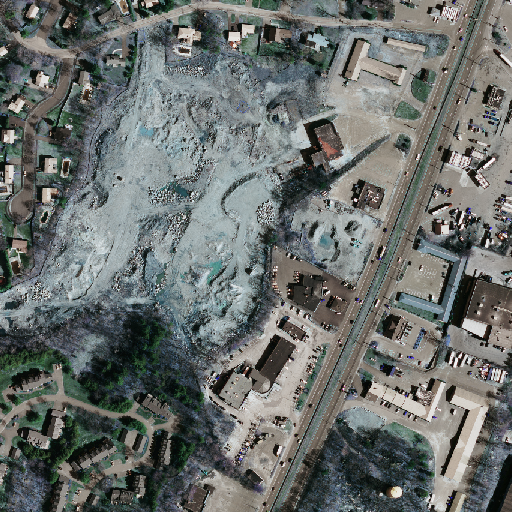} &
        \includegraphics[width=0.15\textwidth]{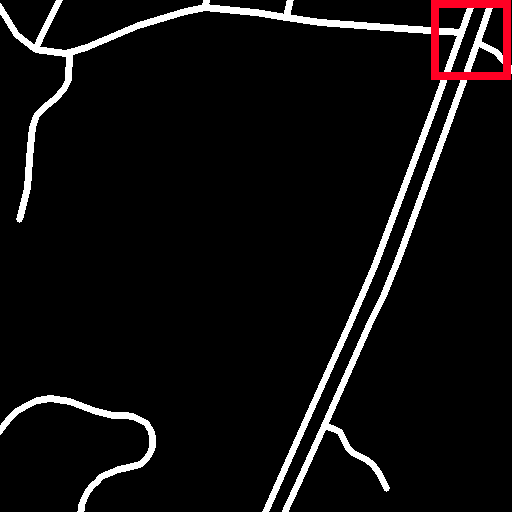} &
        \includegraphics[width=0.15\textwidth]{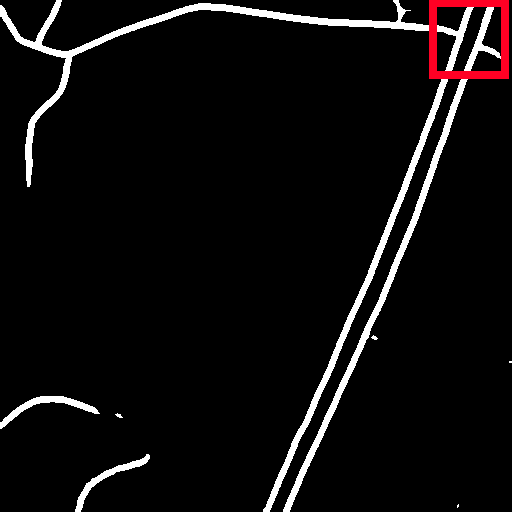} &
        \includegraphics[width=0.15\textwidth]{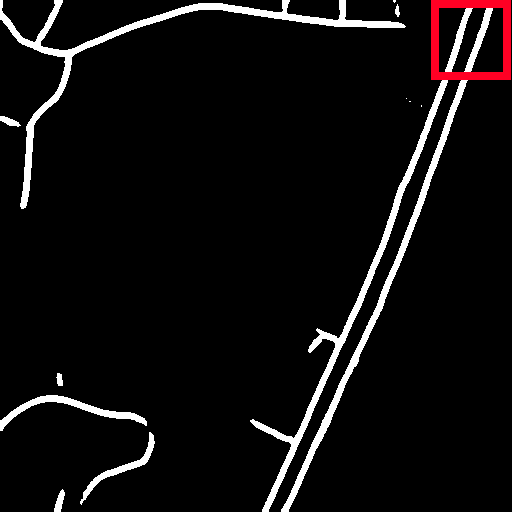}&
        \includegraphics[width=0.15\textwidth]{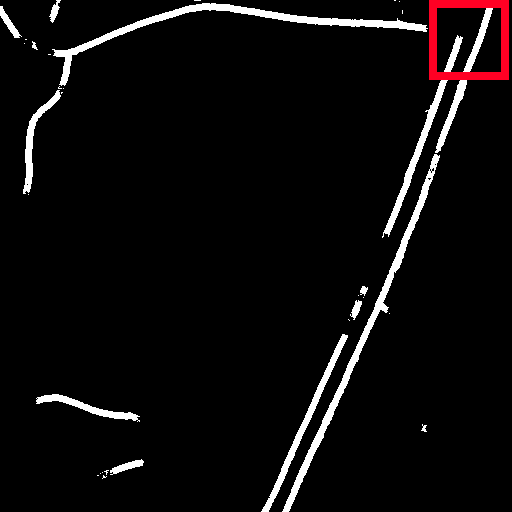} &
        \includegraphics[width=0.15\textwidth]{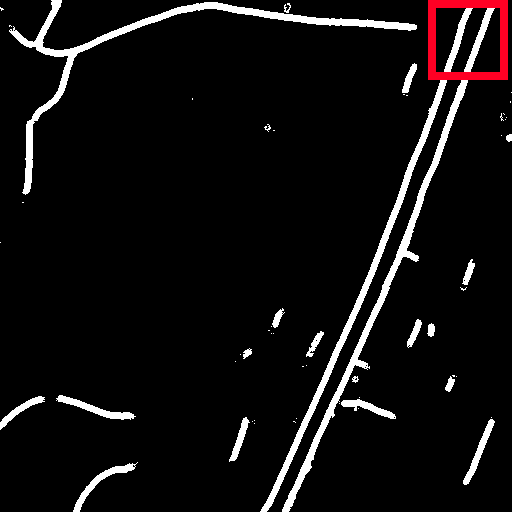} 

        \\
        
        \vspace{2pt}  
        \includegraphics[width=0.15\textwidth]{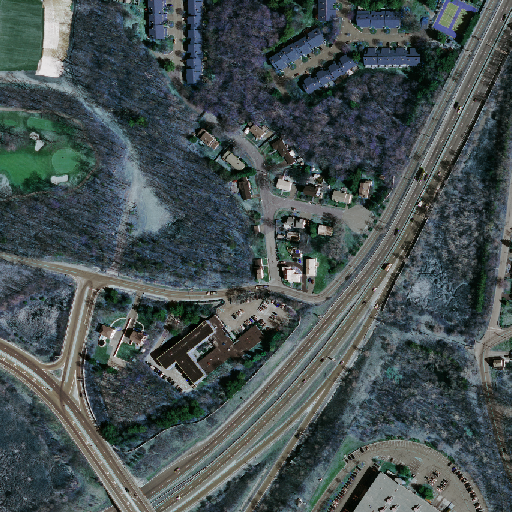} &
        \includegraphics[width=0.15\textwidth]{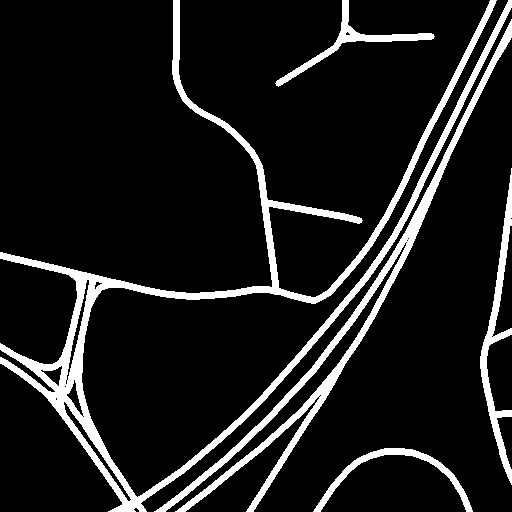} &
        \includegraphics[width=0.15\textwidth]{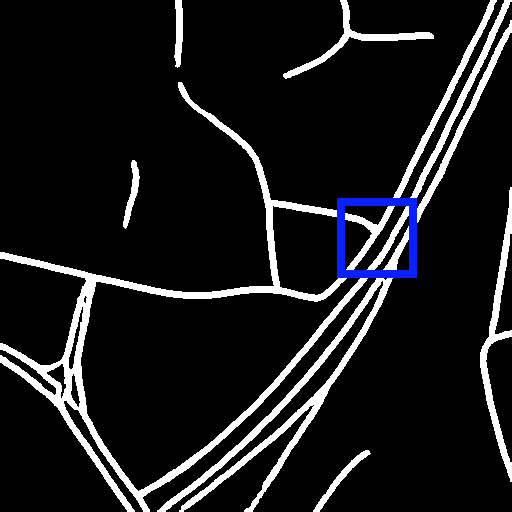} &
        \includegraphics[width=0.15\textwidth]{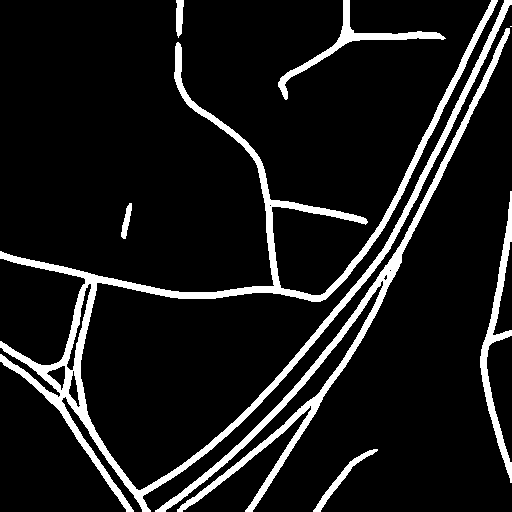} &
        \includegraphics[width=0.15\textwidth]{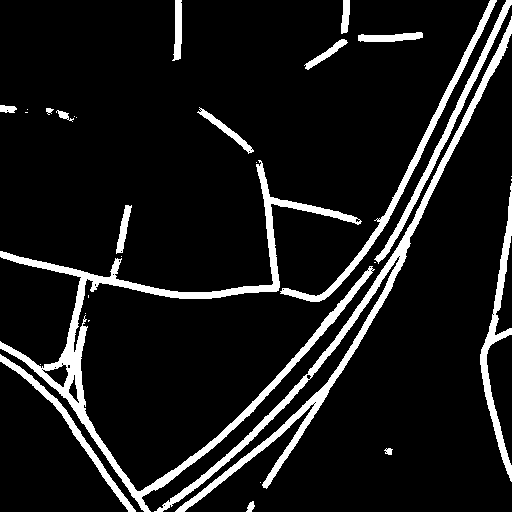} &
        \includegraphics[width=0.15\textwidth]{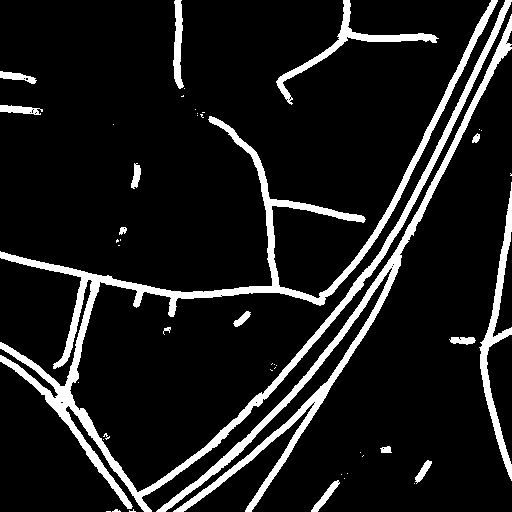}

        \\
        
        \vspace{2pt}  
        \includegraphics[width=0.15\textwidth]{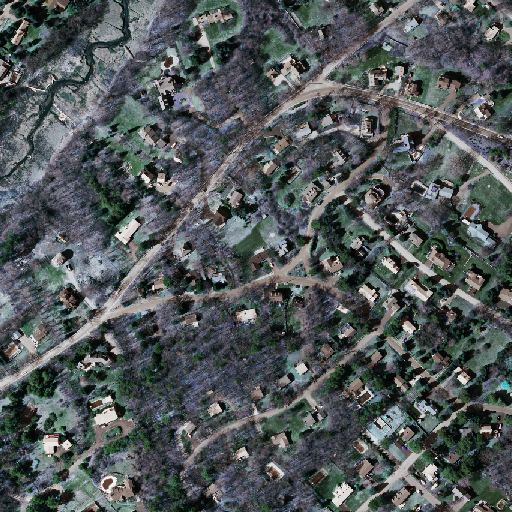} &
        \includegraphics[width=0.15\textwidth]{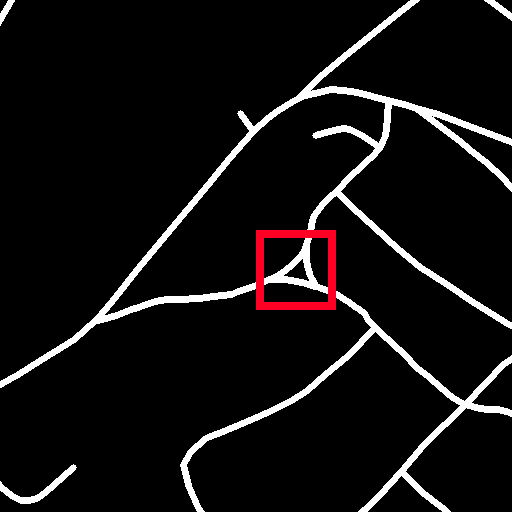} &
        \includegraphics[width=0.15\textwidth]{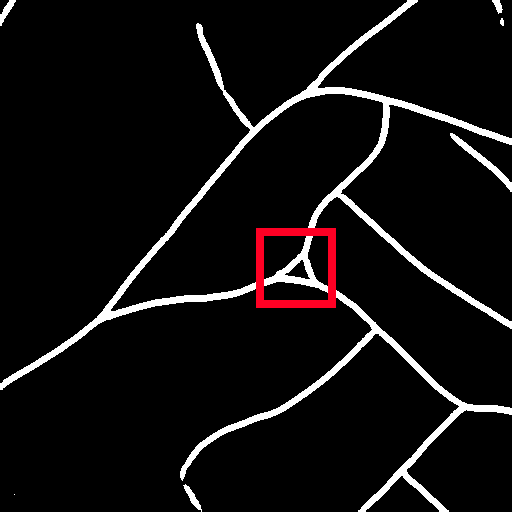} &
        \includegraphics[width=0.15\textwidth]{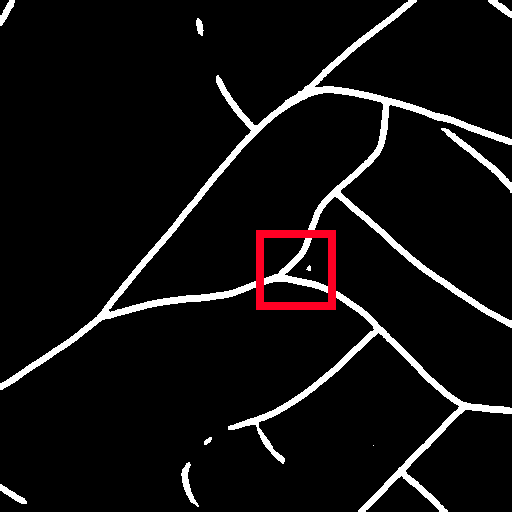}&
        \includegraphics[width=0.15\textwidth]{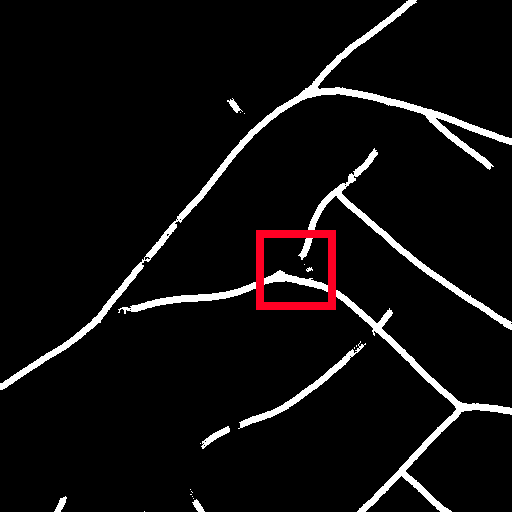} &
        \includegraphics[width=0.15\textwidth]{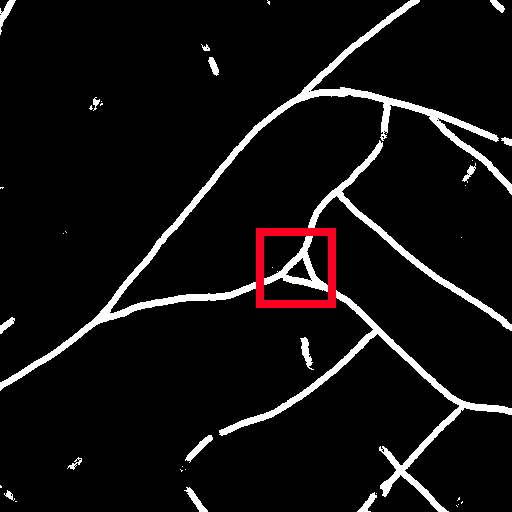}

        \\
        \vspace{2pt}  
        \includegraphics[width=0.15\textwidth]{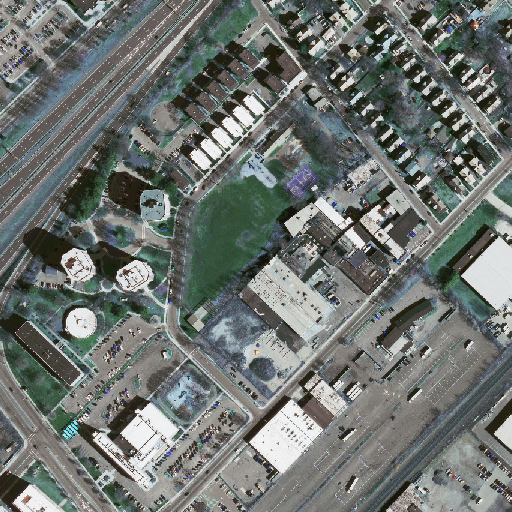} &
        \includegraphics[width=0.15\textwidth]{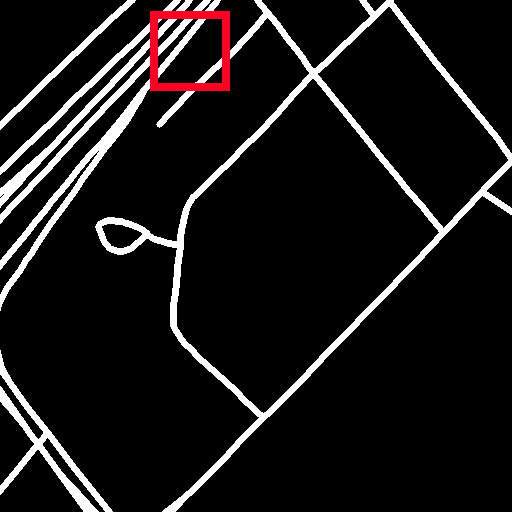} &
        \includegraphics[width=0.15\textwidth]{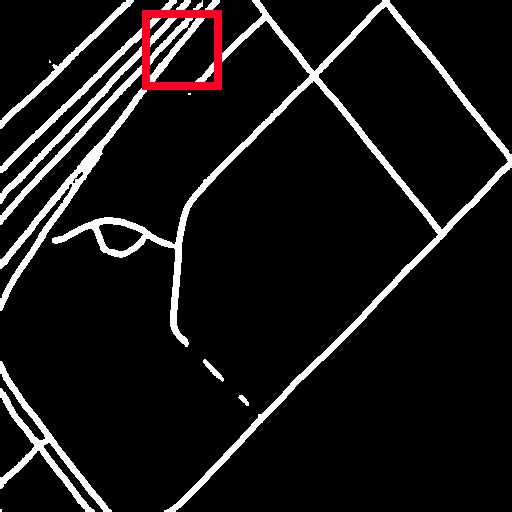} &
        \includegraphics[width=0.15\textwidth]{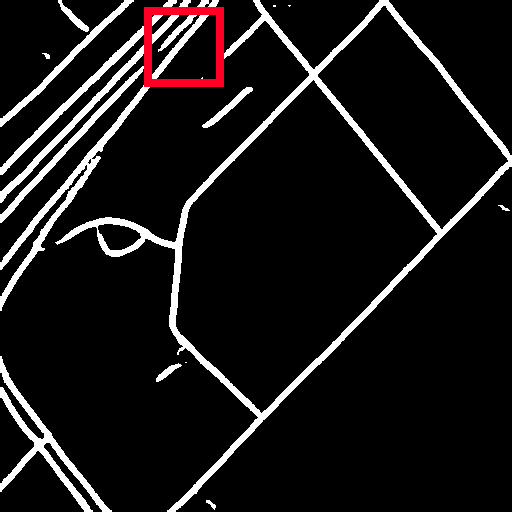}&
        \includegraphics[width=0.15\textwidth]{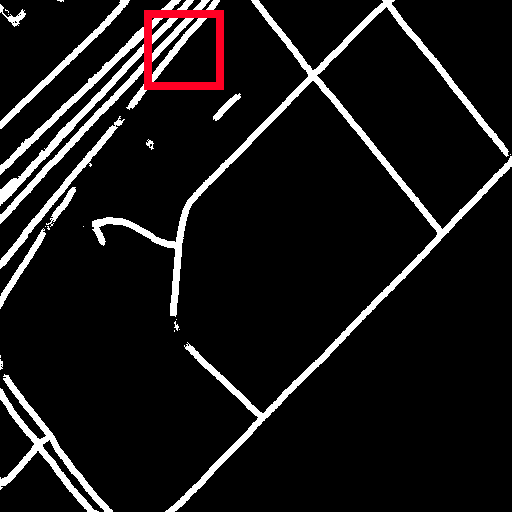} &
        \includegraphics[width=0.15\textwidth]{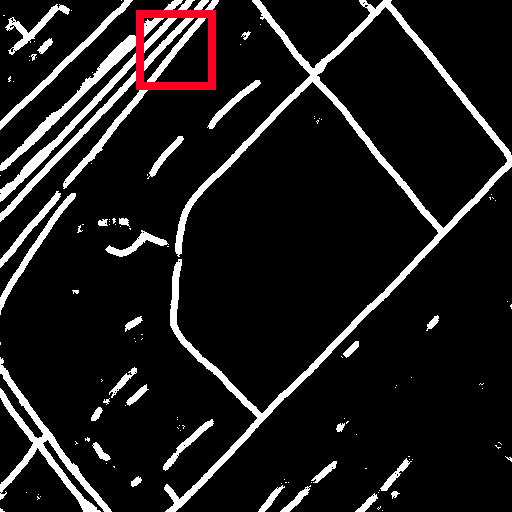}

    \end{tabular}
    \caption{\textbf{Comparative Results of Road Extraction from Massachusetts Dataset}. The figure presents a side-by-side comparison of road extracted by various models, including the Proposed Model, against the Ground Truth, highlighting the effectiveness of each approach in synthesizing accurate road extraction.}
    \label{fig:Massachusettsimage-comparison}
\end{figure*}

\begin{figure*}[!t]
    \centering
    \setlength{\tabcolsep}{3pt}
    \begin{tabular}{cccccc}
        Input & GroundTruth & Proposed & DeepLabV3+ & UNet & SegNet\\
        
        \includegraphics[width=0.15\textwidth]{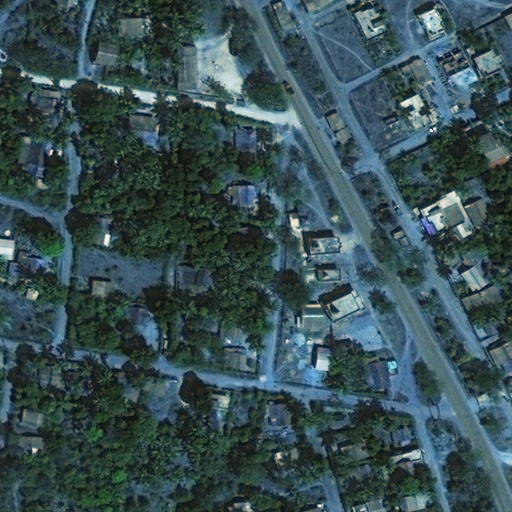} &
        \includegraphics[width=0.15\textwidth]{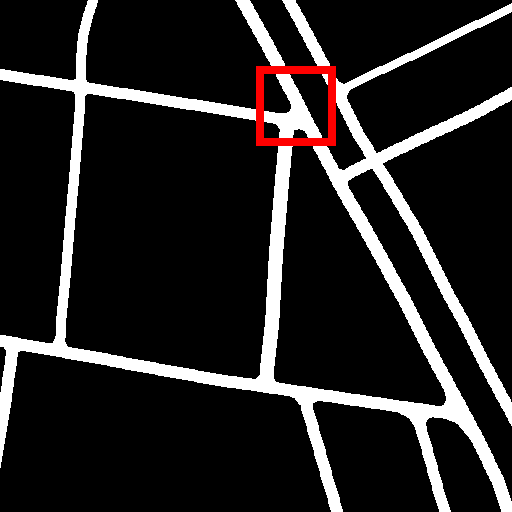} &
        \includegraphics[width=0.15\textwidth]{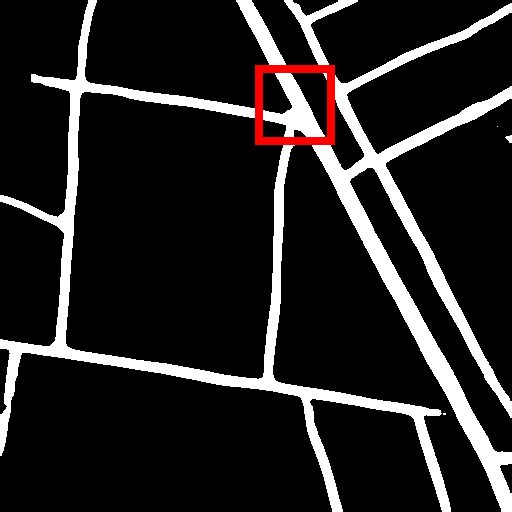} &
        \includegraphics[width=0.15\textwidth]{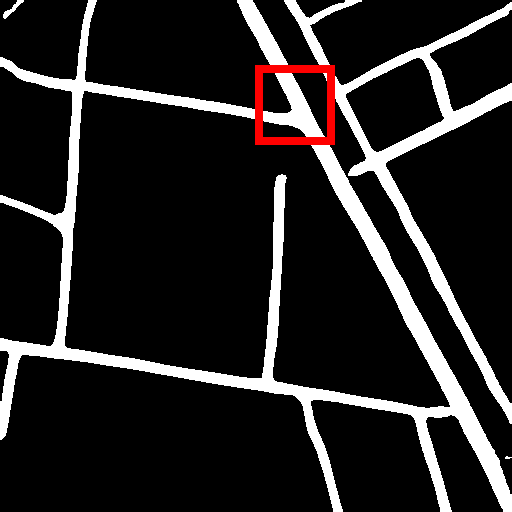} &
        \includegraphics[width=0.15\textwidth]{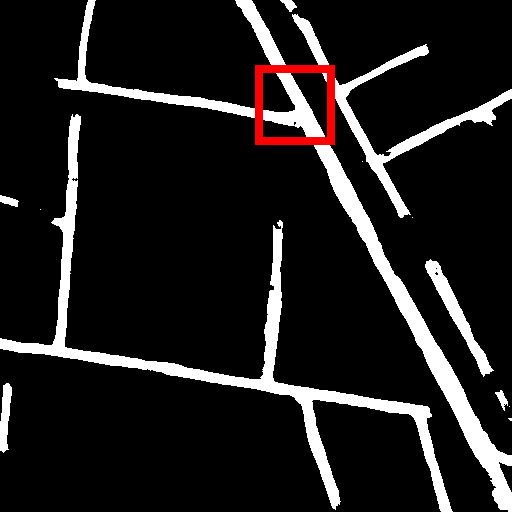} &
        \includegraphics[width=0.15\textwidth]{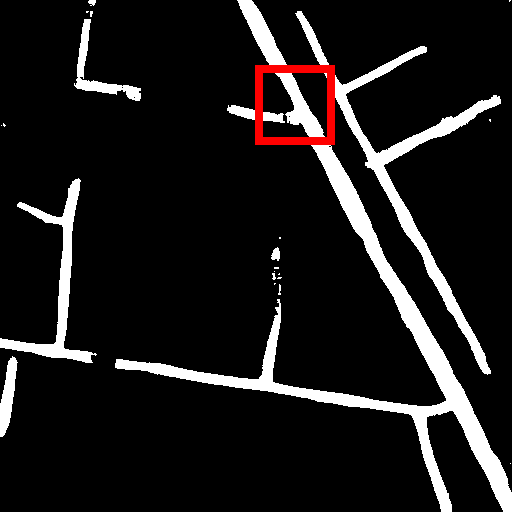} 

        \\
        
        \vspace{2pt}  
        \includegraphics[width=0.15\textwidth]{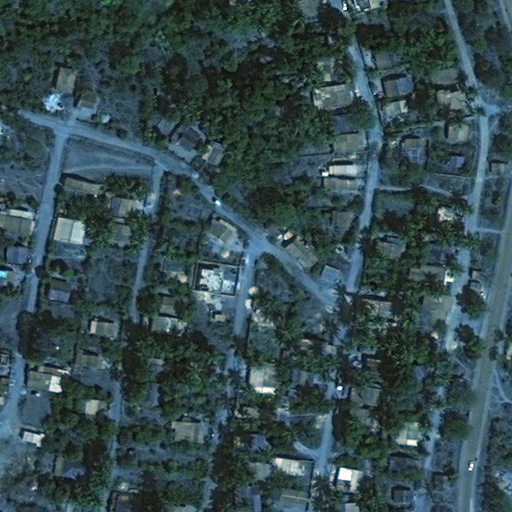} &
        \includegraphics[width=0.15\textwidth]{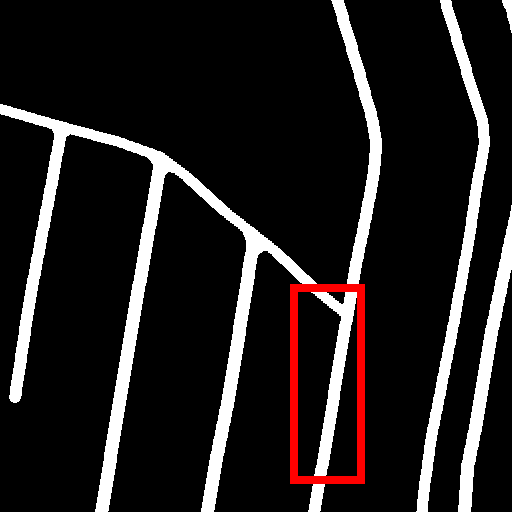} &
        \includegraphics[width=0.15\textwidth]{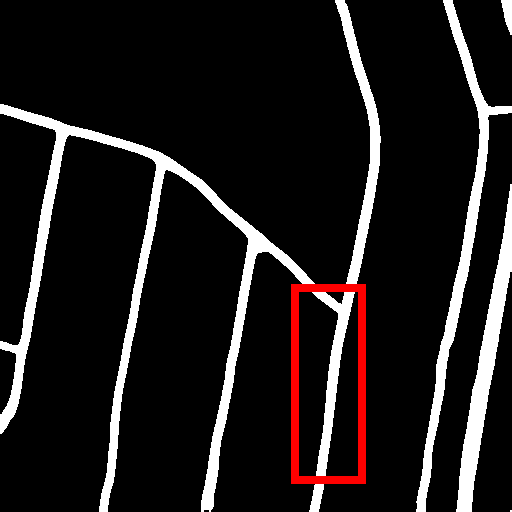} &
        \includegraphics[width=0.15\textwidth]{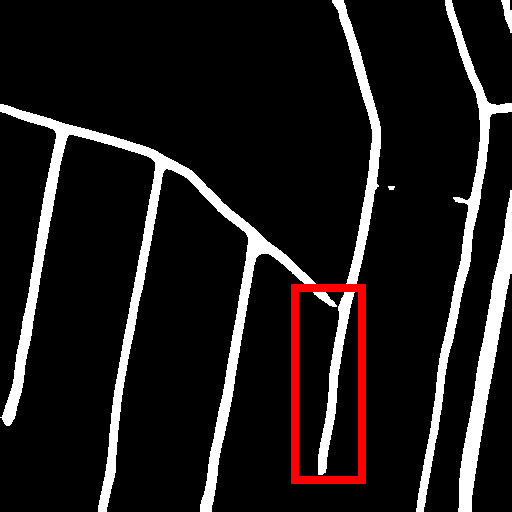} &
        \includegraphics[width=0.15\textwidth]{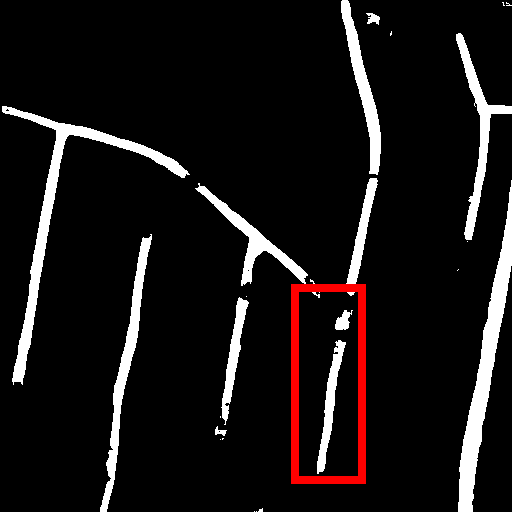} &
        \includegraphics[width=0.15\textwidth]{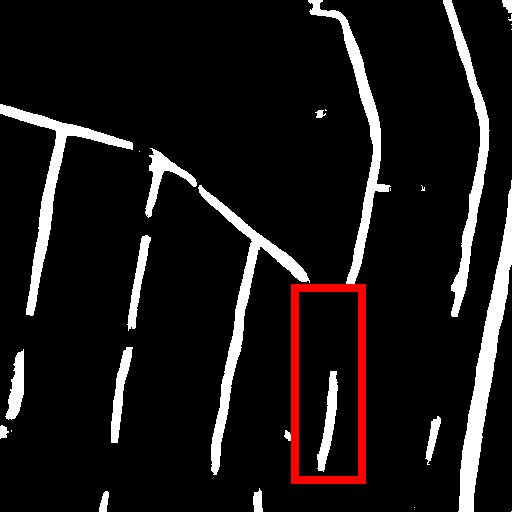}

        \\
        
         \vspace{2pt}  
        \includegraphics[width=0.15\textwidth]{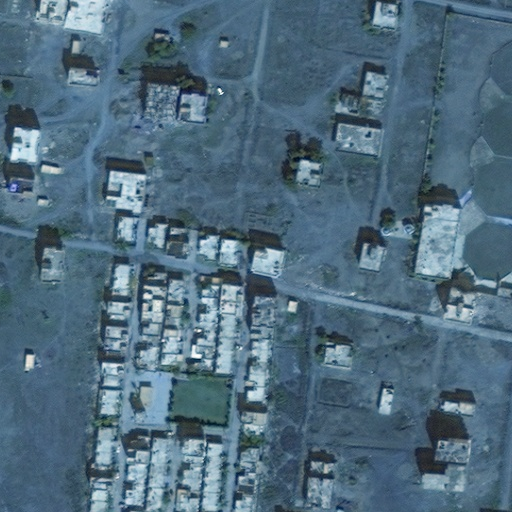} &
        \includegraphics[width=0.15\textwidth]{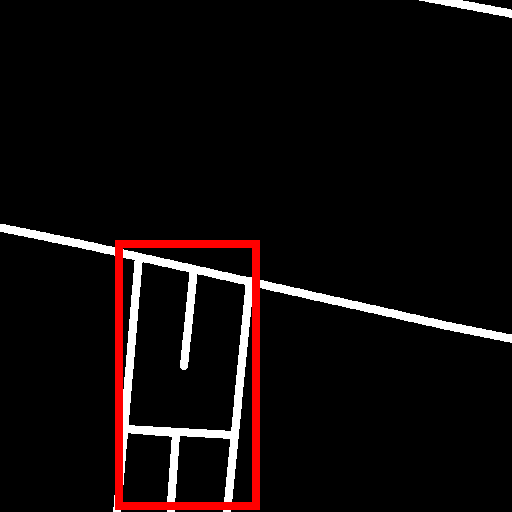} &
        \includegraphics[width=0.15\textwidth]{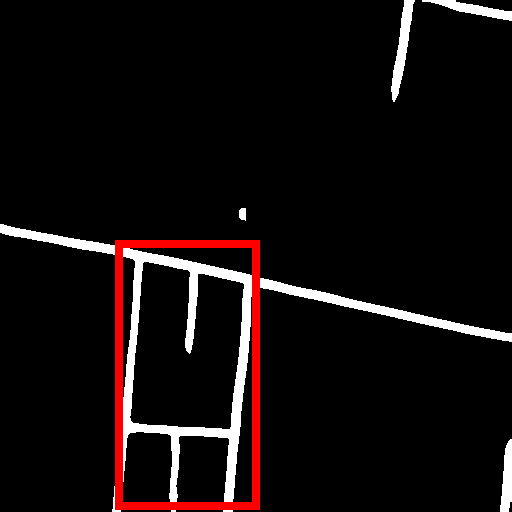} &
        \includegraphics[width=0.15\textwidth]{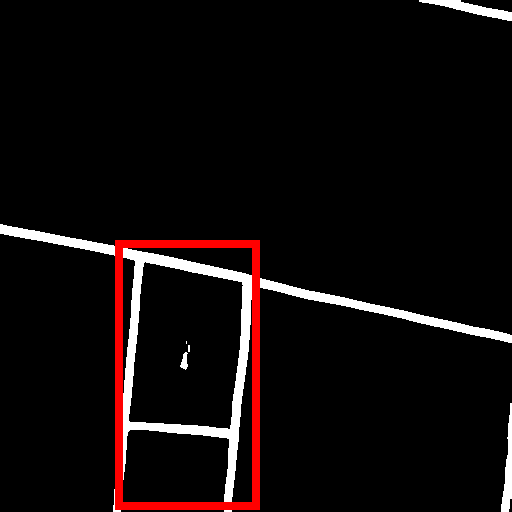} &
        \includegraphics[width=0.15\textwidth]{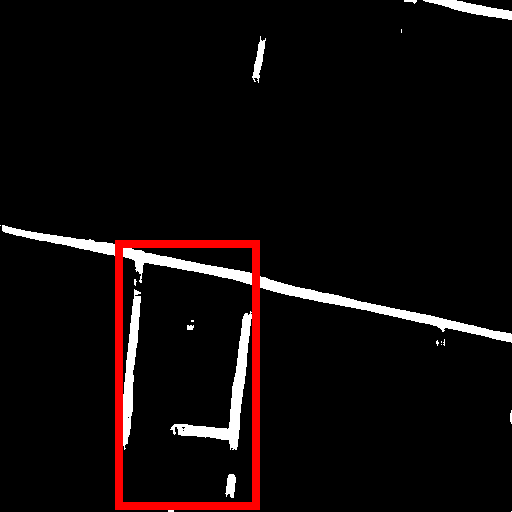} &
        \includegraphics[width=0.15\textwidth]{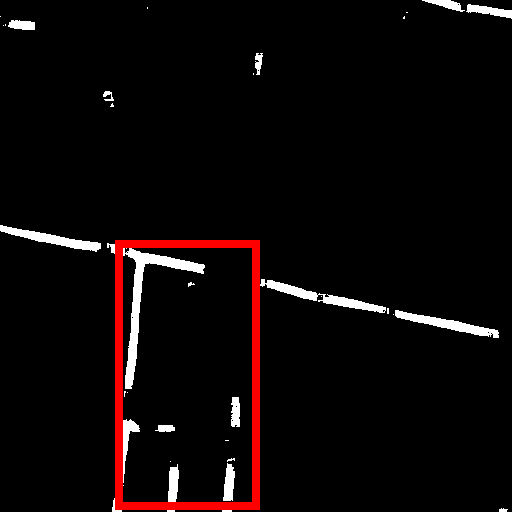}

        \\
        
        \vspace{2pt}  
        \includegraphics[width=0.15\textwidth]{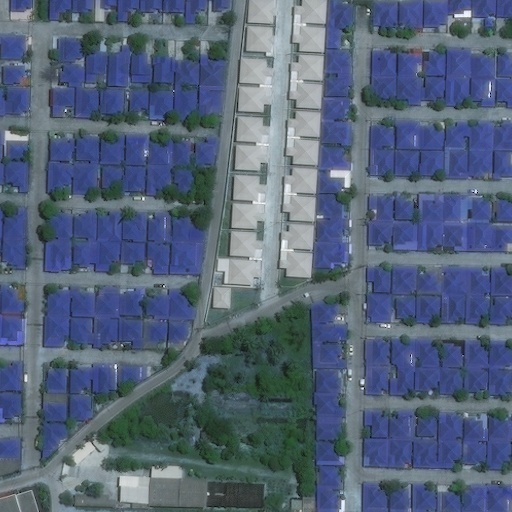} &
        \includegraphics[width=0.15\textwidth]{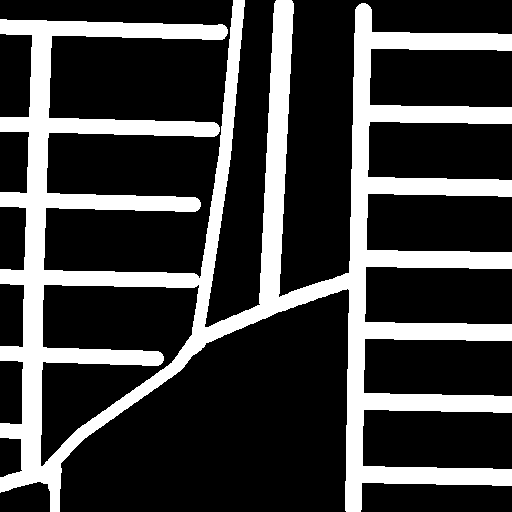} &
        \includegraphics[width=0.15\textwidth]{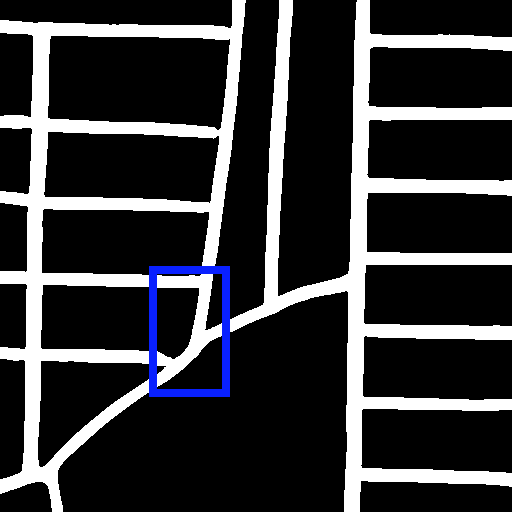} &
        \includegraphics[width=0.15\textwidth]{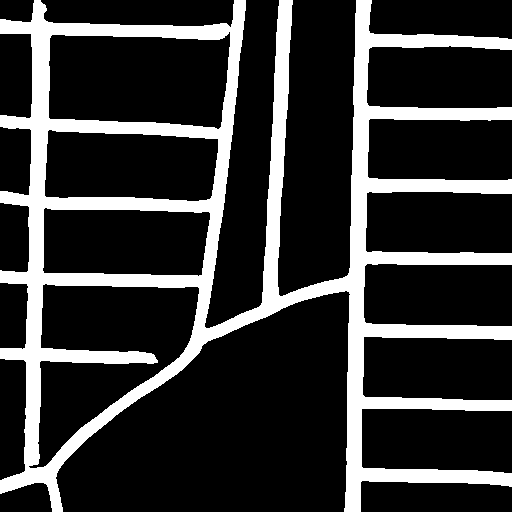} &
        \includegraphics[width=0.15\textwidth]{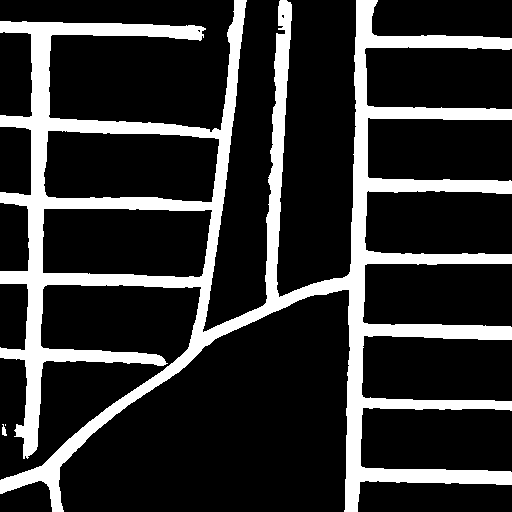} &
        \includegraphics[width=0.15\textwidth]{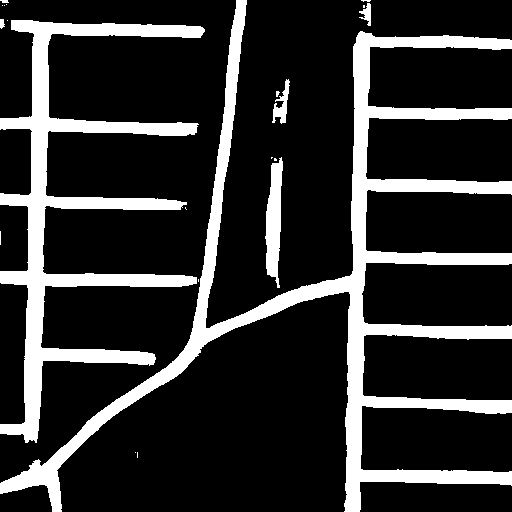} 

    \end{tabular}
    \caption{\textbf{Comparative Results of Road Extraction from DeepGlobe Road Dataset}. The figure presents a side-by-side comparison of road extracted by various models, including the Proposed Model, against the Ground Truth.}
    \label{fig:deepglobeimage-comparison}
\end{figure*}

\section{Discussion}
In addition to the quantitative metrics evaluation, our study also presents visual road extraction depictions of our model, including comparisons with other models. Figure \ref{fig:Massachusettsimage-comparison} presents the road extraction performed on the validation set of the Massachusetts dataset. As depicted in the diagram, our proposed model achieves accurate road extraction in several parts of the images, as highlighted with red-colored squares, compared to most of the other models.

Similarly, our model demonstrates comparable performance on the DeepGlobe road dataset, as shown in Figure \ref{fig:deepglobeimage-comparison}. Accurate road construction and connection can be seen in the red-colored squares in the diagram. An interesting observation is that our model tends to connect roads when there are trees present, as highlighted with blue-colored squares in the second row of Figure \ref{fig:Massachusettsimage-comparison} and the fourth row of Figure \ref{fig:deepglobeimage-comparison}. In the ground truth images, these road connections are not visible. Upon closer inspection, it can be assumed that there is a road occluded by trees, and our model may be able to achieve correct road extraction even in such occluded scenarios.

Based on the quantitative and visualization results, our model outperforms various state-of-the-art models in road extraction. Our study not only demonstrates better performance but also presents an efficient computational approach by adopting the depthwise separable mechanism and increasing the field-of-view with the dilation mechanism. To provide a more intuitive and mathematical understanding of how depthwise separable dilated convolution used in our work brought efficiency compared to standard convolution, we can express the computational savings mathematically.

Given an image of size $512 \times 512 \times 3$ and using a $3 \times 3$ kernel, the standard convolution operation can be expressed as:
\begin{equation*}
\text{Operations}_{\text{standard}} = H \times W \times K \times K \times C_{\text{in}} \times C_{\text{out}}
\end{equation*}

Where: $H = 512$ (height of the image), $W = 512$ (width of the image), $K = 3$ (size of the kernel), $C_{\text{in}} = 3$ (number of input channels) and $C_{\text{out}} = 64$ (number of output channels). Substituting these values, we get:
\begin{equation*}
\text{Operations}_{\text{standard}} = 512 \times 512 \times 3 \times 3 \times 3 \times 64 = 151,165,440
\end{equation*}

Since depthwise separable convolution involves two steps, the operations can be split into two parts: depthwise convolution and pointwise convolution, and their operations are expressed below as:
\begin{equation*}
\text{Operations}_{\text{depthwise}} = H \times W \times K \times K \times C_{\text{in}}
\end{equation*}
\begin{equation*}
\text{Operations}_{\text{pointwise}} = H \times W \times C_{\text{in}} \times C_{\text{out}}
\end{equation*}

Combining both depthwise and pointwise operations, and substituting all the values, the total number of operations for depthwise separable convolution is:
\begin{equation*}
\text{Operations}_{\text{depthwise separable}} = 7,077,888 + 50,331,648 = 57,409,536
\end{equation*}

To compare the efficiency, we can take the ratio of the operations required:
\begin{equation*}
\frac{\text{Operations}_{\text{depthwise separable}}}{\text{Operations}_{\text{standard}}} = \frac{57,409,536}{151,165,440} \approx 0.38
\end{equation*}

This indicates that the depthwise separable convolution is approximately 62\% more efficient than standard convolution based on operations.

Furthermore, by incorporating the dilation mechanism in depthwise convolution, the receptive field is increased without adding extra parameters, enhancing the network's ability to capture multi-scale features more effectively. The dilation rate $r$ introduces spacing between kernel elements, allowing for a larger coverage area without increasing the kernel size.

In summary, the utilization of depthwise separable dilated convolution in our DenseDDSSPP module brings both computational efficiency and enhanced feature extraction capabilities, as evidenced by the better performance of our proposed model in road extraction tasks.

\section{Conclusion}
Road extraction is one of the most important research areas for applications such as autonomous navigation and smart city planning, yet it faces several challenges. Our study addresses these challenges by advancing the capabilities of the DeepLabV3+ model through the introduction of the Dense Depthwise Dilated Separable Spatial Pyramid Pooling (DenseDDSSPP) module, replacing the standard ASPP module. The study also identifies Xception as the optimal backbone network for enhanced feature extraction and integrates the Squeeze-and-Excitation block into the decoder, facilitating channel-wise learning to emphasize relevant features. To the best of our knowledge, our work is the first attempt to incorporate Dense Depthwise Dilated Separable Convolution to form a DenseDDSSPP module within a semantic segmentation model for road extraction from satellite imagery. The proposed approach has demonstrated better performance on publicly available datasets such as the Massachusetts road dataset and the DeepGlobe road dataset. Our model outperformed several state-of-the-art models based on various evaluation metrics in a supervised learning setup, improving IoU, precision, and \(F_1\) score metrics. The visual comparisons further highlighted our model’s ability to accurately extract and connect road segments, even in scenarios where roads are occluded by trees.

Future work will focus on enhancing the generalizability of our model by collecting additional datasets and evaluating its performance in zero-shot scenarios \cite{bucher2019zero}, enabling the semantic segmentation of road types not encountered during training. We also aim to transition from a supervised setup, which requires annotated images for training, to a self-supervised learning approach. Inspired by recent advancements in self-supervised learning, such as those by Hou et al. \cite{hou2021c} and Mahara et al. \cite{mahara2024generative}, we plan to develop methods for road extraction that do not rely on corresponding annotated images for each satellite image. This approach holds the potential to significantly broaden the applicability of our model to diverse and unannotated datasets.

\section*{Acknowledgment}
This material is based in part upon work supported by the National Science Foundation under Grant Nos. CNS-2018611 and CNS-1920182 and by Florida Department of Environmental Protection Grant C-2104

\nocite{*}
\bibliographystyle{IEEEtran}

\end{document}